\documentclass{ametsocV6.1}
\usepackage{color,soul}
\usepackage{amsmath}
\usepackage[normalem]{ulem}
\usepackage{xcolor}
\usepackage{graphicx}
\usepackage{subcaption} 
\usepackage{adjustbox}
\usepackage{amsfonts} 
\AtBeginDocument{%
  \nolinenumbers
  \let\internallinenumbers\relax
}

\title{Probabilistic adjustment of seasonal forecasts using cVAE with a non-parametric decoder: A case study of Arctic sea ice predictions}

\authors{Parsa Gooya$^{1}$\correspondingauthor{Parsa Gooya, parsa.gooya@ec.gc.ca} and Reinel Sospedra-Alfonso$^{2}$ } \affiliation{ $^{1}$ Canadian Centre for Climate Modelling and Analysis, Environment and Climate Change Canada, Victoria, British Columbia, Canada\\ $^{2}$ Climate Modelling Research Division, Environment and Climate Change Canada, Victoria, British Columbia, Canada}

\abstract{Seasonal climate predictions support planning and risk management by offering early information of the likely climate conditions in the coming months, and associated uncertainties. Ensemble forecasts enable this by simulating many plausible outcomes, allowing predictions to be expressed as usable probabilities. Large ensembles and high-resolution forecasts strengthen this by better sampling uncertainty and capturing finer-scale processes but come with significant computational cost. Moreover, forecast ensembles drift and exhibit systematic biases and spatio-temporal errors that grow with lead time, requiring careful post-processing and calibration. A probabilistic post-processing framework based on conditional Variational Autoencoders (cVAEs) was developed at the Canadian Center for Climate Modeling and Analysis to generate large ensembles of adjusted seasonal predictions of Arctic sea ice. The generative model was designed to learn the observational distribution conditioned on the biased model prediction. This enables generation of arbitrarily large ensembles of calibrated, bias corrected forecasts with improved skill. Here, we extend this framework to a Generalized Variational Inference problem to address the loss of fine-scale energy and the characteristic blurriness, a known limitation of standard cVAEs. Specifically, we employ a generator in place of the Gaussian parametrized decoder in cVAE and use Continuous Ranked Probability Score (CRPS) in the objective function instead of the Gaussian log-likelihood. We further use a higher resolution target dataset compared to the raw forecast. We show that the adjusted forecasts are better calibrated, more consistent with the observational distribution, and exhibit smaller errors than benchmark predictions, while also enhancing the resolution of the raw forecasts and improving sharpness and spectral power relative to the standard cVAE.}

\begin{document} 
\maketitle

\section{Introduction}

\noindent Seasonal predictions refer to forecasts on time scales ranging from a few months up to a year. In Canada, seasonal forecasts of key climate variables are produced by the Canadian Seasonal to Interannual Prediction System (CanSIPS; \citealp{canSIPS_2013, CanSIPsv2_2020}), which combines two fully coupled ocean-atmosphere-land-sea ice climate models. Each month, the models are initialized using estimates of the current state of the climate and then integrated forward in time for up to 12 months. In the third generation of the Canadian prediction system (CanSIPSv3), each model produces 20 ensemble members, 10 initialized on the first of the month, and 10 initialized 5 days before, to form the multi-model ensemble forecasts. One of the contributing models is the Canadian Earth System Model version 5.1 \cite[CanESM5.1,][]{saadgkmrssvaclllmrssssswy23}, developed at the Canadian Centre for Climate Modelling and Analysis (CCCma) and derived from CanESM5.0 \citep{Swart-2019}. \\

\noindent Successive versions of CanSIPS have demonstrated skill in a range of variables, including those related to the cryosphere \citep{sigmond2013, sigmond2016, smk16, smmd16,  smklldm24, Dirkson2021}. Such variables tend to draw skill from their ``memory" or persistence of their initial anomalies, and from interactions with local or remote drivers. However, inherent model deficiencies and limitations of the initialization process lead to complex, often nonlinear systematic errors that grow with integration time (or lead month, defined as the number of months after initialization for which a forecast is issued). These deficiencies impact forecast skill, hampering CanSIPS' ability to capitalize on key sources of predictability. To mitigate such biases, post-processing techniques are routinely applied, with methods that range from simple lead time dependent climatological bias corrections to more sophisticated machine learning ((ML)) and non-ML methods. \\ 

\noindent For short- to medium-range weather prediction, machine learning and, in particular deep learning, has been implemented to improve the aggregate output statistics of prediction systems, including the spread of an ensemble of deterministic forecasts \citep{sacco:hal-03685523}, corrections to probabilistic forecasts \citep{DLpost-processing}, and artificial generation of high-dimensional weather samples from the target forecast distribution using generative models \citep{diffensemblegen2024}. For decadal climate predictions, which extend the seasonal forecast range for up to a decade, lead time dependent linear trend corrections  \citep{kbmsl12} and other post-processing methods with various levels of complexity \citep{pblrmu18,nvwu19,Kharin2017} have been proposed, including deep learning-based adjustments  \citep{Sospedra_Alfonso_2024}. \\

\noindent Among cryospheric variables with proven forecasting skill in CanSIPS is sea ice concentration (SIC) \citep[e.g.,][]{sigmond2013,sigmond2016}, defined as the fraction of sea ice within a model grid cell. CanSIPS' Arctic SIC forecasts and their downstream products have contributed to several multi-model efforts and outlooks, including the Arctic Climate Forum, a core activity of the World Meteorological Organization's (WMO's) Arctic Regional Climate Centre Network (\url{https://www.arctic-rcc.org/acf}). Such forecasting products are important because Arctic sea ice cover is a key regulator of the polar climate system, shaping albedo feedbacks, ocean-atmosphere heat exchange, and large-scale circulation. Moreover, the rapid decline of Arctic sea ice cover \citep{serreze2007, Cavalieri_Parkinson_2012}, especially in summer, poses a threat to local communities and ecosystems, but also creates economic opportunities for marine fishing, shipping, tourism and resource extraction \citep{Sea_Ice_Initialisation_On_Seasonal_Prediction_Skill, SubseasonaltoSeasonalArcticSeaIceForecastSkillImprovementfromSeaIceConcentrationAssimilation}. Therefore, preparing, mitigating and planning in response to such changes and their impacts requires accurate and reliable seasonal and longer term predictions \citep{Wagner02072020}. \\

\noindent For seasonal SIC prediction, common post-processing methods include simple lead dependent climatological bias corrections, linear regression approaches \citep{Blanchard-Wrigglesworth2017}, statistical calibration methods based on parametric distribution estimation \citep{CalibrationSubseasonalSICForecasts, CalibratedProbabilisticForecastsofArcticSeaIceConcentration}, and more recent machine learning models \citep{Palerme2024, He2025MLICE}. These techniques rely either on deterministic corrections, i.e. are constrained by the small number of computationally expensive members of the raw forecast ensemble, or assume local parametric distribution forms and estimate their parameters based on a limited forecast ensemble. Furthermore, most non-ML methods apply transformations to disjoint grid-level marginal distributions, in some cases extended to include local information to estimate the parametric distribution but applied nevertheless at grid-cell level \citep{CalibrationSubseasonalSICForecasts}. As a result, even if they are statistically calibrated, these forecasts can struggle in generating large ensembles of high resolution maps that represent the spatial structure of extremes. Large ensembles of fully resolved maps can overcome this by sampling many plausible climate realizations and produce reliable probabilistic forecasts, stronger diagnostics of model biases, clearer quantification of internal variability, physically meaningful spatial patterns and better representation of uncertainty, including improved sampling of low-probability, high-impact extremes. \\

\noindent \cite{gooya2025probabilisticbiasadjustmentseasonal}, hereafter GSA2025, proposed a probabilistic adjustment framework built on a generative ML model for seasonal predictions of Arctic SIC aiming at overcoming these limitations. The framework relies on a conditional Variational Autoencoder (cVAE) model tasked with learning the distribution of observations conditioned on the biased ensemble mean of the raw seasonal forecasts. With this approach, an arbitrary large ensemble of adjusted forecasts is produced. However, while this approach improves the issues discussed above, it has drawbacks: adjusted forecasts are overly smooth and have muted variability at fine spatial scales, a known issue with the original cVAE formulation \citep{ VAEGAN2016, xu2025likelihoodfreevariationalautoencoders, dorta2018structureduncertaintypredictionnetworks}. \\

\noindent The present study extends the work in GSA2025 to address these shortcomings, in particular the muted variability and ``blurry" appearance of standard cVAE models, where fine-scale gradients, edges, and localized spatial variations are washed out by the model's tendency to output the most likely outcome out of the possible ones. We also examine the potential for this probabilistic framework to produce corrected forecasts at higher spatial resolution, making it potentially a post-processing and downscaling tool. The remainder of the paper is organized as follows. Section \ref{sec:methods} and Appendix \ref{sec:app_cVAE} detail the cVAE framework, its formulation for adjustment of seasonal predictions, the cVAE-CRPS model, the architecture of the neural network, and the training/inference procedures. The data used for training and evaluation are described in section \ref{sec:data}. Probabilistic and deterministic evaluation metrics are introduced in section \ref{sec:evaluation_metrics} and Appendix \ref{sec:app_metrics}, and the results are presented in section \ref{sec:results}. Discussions and conclusions are provided in section \ref{sec:discussions_and_conclusions}.

\section{Methods}
\label{sec:methods}

\subsection{Conditional Variational Autoencoder (cVAE)}
\label{subsec:cVAE}

\noindent The generative model  of  
GSA2025 is based on a conditional Variational Autoencoder \citep[cVAE,][]{cVAE2015} that models the target variable ${y}\in\mathcal{Y}$ with the help of an unobservable (latent) variable ${z}\in \mathcal{Z}$ \citep{cVAE2015, prince2023understanding}, conditioned on the input variable $x\in\mathcal{X}$. The latent variable $z$ can be regarded as a lower-dimensional representation of ${y}$ that explains the variations in the data in a condensed manner. The generative (inference) process uses samples in the latent space $\mathcal{Z}$ drawn from the conditional prior distribution $p_{\omega}({z|x})$, which is generally assumed to be Gaussian dependent on the condition. The generated data ${y} \in \mathcal{Y}$ are sampled from the distribution $p_{\theta}({y} | z, x)$, referred to as probabilistic $\textit{decoder}$. In a standard cVAE model, the decoder is assumed to be a Gaussian distribution \citep{cVAE2015}, but we follow a different approach here (section \ref{sec:methods}\ref{subsec:cVAE-CRPS}).\\




\noindent Distribution parameters (e.g., mean and standard deviation of the decoder distribution and the conditional prior) are learned using maximum likelihood estimation, which optimizes the parameters $\theta$ and $\omega$ of the neural network models to maximize the log-likelihood of the observed targets in the generative output distribution \citep{cVAE2015, prince2023understanding}. Kingma \& Welling (2014) showed that the VAE parameters can be efficiently estimated using the variational lower bound of the log-likelihood as a surrogate objective function. For cVAE, 
\begin{equation}
     \log p_{\theta,\omega}(y|x) = \mathrm{KL} \left( q_{\phi}({z} | y,x) \Vert p_{\theta,\omega}({z} | y, x) \right) +  \mathcal{L}(y,x;\theta,\phi,\omega)  \geq  \mathcal{L}(y,x;\theta,\phi,\omega)
     \label{eq:log_p(x|y)}
\end{equation}
where $\log p_{\theta,\omega}(y|x)$ is the generative (inference) distribution under the model and the surrogate objective function $ \mathcal{L}(y,x;\theta,\phi,\omega)$ is the Evidence Lower Bound (ELBO) 
\begin{equation}
    \mathcal{L}(y,x;\theta,\phi,\omega) = \mathbb{E}_{q_{\phi}({z} | y, x)} \left[ \log p_{\theta}({y} | z, x) \right] -  \mathrm{KL} \left( q_{\phi}({z} | y, x) \Vert p_{\omega}(z|x) \right)
    \label{eq:elbo}
\end{equation}
and $\mathrm{KL}(q\Vert p)$ is the Kullback-Leibler divergence, which is always non-negative and vanishes when $q=p$ almost everywhere. This implies that the lower bound in Eq. \ref{eq:log_p(x|y)} is tight when $q_{\phi}({z} | y,x) = p_{\theta,\omega}({z} | y, x)$. The distribution $q_{\phi}({z} | y, x)$, known as the probabilistic \textit{encoder} and also assumed to be Gaussian, is thus used to approximate the true intractable posterior $p_{\theta,\omega}({z} | y, x)$ \citep{kingma2014, cVAE2015}. Its parameters are learned during the training process through a separate neural network with parameters $\phi$. At population-level optimum when ELBO is tight and under sufficient model capacity, the resulting generative distribution ($p_{\theta,\omega}(y\mid x)$) recovers the true data-generating distribution ($p_D(y\mid x)$;  \citealp[]{yacoby2020characterizing}).

\subsection{Probabilistic adjustment of seasonal predictions}
\label{subsec:prob_Badj}

\noindent For probabilistic adjustment of seasonal prediction ensembles, the goal is to learn a mapping from a biased \textit{ensemble mean} forecast $\bar x_{tl}$, where $t$ denotes the initialization time and $l$ the lead month, to the observational distribution $p(Y|\bar x_{tl})$ (GSA2025). Although biased, $\bar x_{tl}$ simulates the predictable component of the climate system due to both external forcing and internal variability, the latter only possible due to the initialization of the forecast. The observation $y_{tl}$, indexed for simplicity in the same initial-time and lead-month notation as the forecast, is the realized outcome of the climate system drawn from a range of possible states, which define the target observational distribution. We are thus interested in modeling the observational distribution conditioned on the biased ensemble mean, which carries the predictable component of the forecasts. GSA2025 achieve this by maximizing the log-likelihood of the observation $y_{tl}$ in the target distribution conditioned on $\bar x_{tl}$ ($\max{_\theta}$  $\log p_{\theta}(Y = y_{tl}| \bar x_{tl})$) with the cVAE model formulated as:

\begin{equation}
    \begin{aligned}
        &\textit{Encoder:} \quad q_{\phi}({z} | {y_{tl}}, {\bar x_{tl}}) = \mathcal{N} \left( {\mu}_{NN_{\phi}}({y_{tl}}, {\bar x_{tl}}), {\sigma}_{NN_{\phi}}^2({y_{tl}}, {\bar x_{tl}}) {I} \right) \\
        &\textit{Decoder:} \quad p_{\theta}({y} | {z}, {\bar x_{tl}}) = \mathcal{N} \left( {\mu}_{NN_{\theta}}({z}, {\bar x_{tl}}), {\sigma^2}{I} \right) \\
        &\textit{Prior:} \qquad p_{\omega}({z} | {\bar x_{tl}}) = \mathcal{N} \left( {\mu}_{NN_{\omega}}({\bar x_{tl}}), {\sigma}_{NN_{\omega}}^2( {\bar x_{tl}}) {I} \right)\\
    \end{aligned}
    \label{eq:cVAE_model}
\end{equation}

\noindent where ${I}$ is the identity matrix and ${\sigma^2}$ is a location independent (constant) decoder noise. Under this formulation, the $\mathrm{KL}$ term in the objective function (Eq. \ref{eq:log_p(x|y)}) has a closed form solution \citep{cVAE2015}, while the second term reduces to a quantity proportional to the Mean Square Error (MSE) between the decoder mean ${\mu}_{NN_{\theta}}({z}, {\bar x_{tl}})$ and the target observation $y_{tl}$. Minimizing this MSE is equivalent to maximizing the observation's log-likelihood under the decoder's Gaussian distribution. Note that the Gaussian formulation of the decoder in Eq. (\ref{eq:cVAE_model}) does not imply Gaussian marginal (grid-wise) SIC distributions since the decoder is conditioned on latent samples $z$. The expectation $\mathbb{E}_{q_{\phi}({z} | y_{tl}, \bar{x}_{tl})} \left[ \log p_{\theta}(Y = y_{tl} | z, \bar{x}_{tl}) \right]$ in Eq. \ref{eq:log_p(x|y)} is further simplified by drawing samples ${z}^{n}_{tl}$ ($n = 1, \dots, N$) using the posterior distribution $q_{\phi}({z} | y_{tl}, \bar{x}_{tl})$. With this formulation, the cVAE objective function simplifies to:

\begin{eqnarray}
\mathcal{L}(x;\theta,\phi) & = & \mathrm{KL} \Bigl(q_{\phi}(z \mid y_{tl}, \bar{x}_{tl}) \,\Vert\, p_{\omega}(z \mid \bar{x}_{tl}) \Bigr) - 
\frac{1}{N} \sum_{n=1}^{N} \log p_{\theta} \bigl( Y =  y_{tl} \mid z^{n}_{tl}, \bar{x}_{tl} \bigr) \nonumber \\
& \propto & \mathrm{KL} \Bigl(q_{\phi}(z \mid y_{tl}, \bar{x}_{tl}) \,\Vert\, p_{\omega}(z \mid \bar{x}_{tl}) \Bigr)  +
\frac{1}{N} \sum_{n=1}^{N}  \mathrm{MSE} \bigl( y_{tl}, \mu_{\mathrm{NN}_\theta} (z^{n}_{tl}, \bar{x}_{tl}) \bigr).
\label{eq:objective_function}
\end{eqnarray} 

\noindent Note that the negative log-likelihood on the right-hand side of Eq. \ref{eq:objective_function} is minimized for \textit{each} ${z}_{tl}^{{n}}$ using MSE. 

\subsection{The cVAE-CRPS model}
\label{subsec:cVAE-CRPS}

\noindent One issue with using the standard formulation of cVAE models, and more generally models with MSE as their objective function, is the loss of spectral energy at fine scales manifested as smooth (blurry) outputs where the high resolution structure and variability is underrepresented \citep{MLclimate2024, VAEGAN2016, xu2025likelihoodfreevariationalautoencoders, VAEGANdownscalingPrecipitation2022, dorta2018structureduncertaintypredictionnetworks}. In cVAE, the important aspects of the data are explained by the latent variable, and the remaining unmodeled aspects are ascribed to the decoder noise \citep{prince2023understanding}. As mentioned above (section \ref{sec:methods}\ref{subsec:prob_Badj}), the MSE in the cVAE loss function results from the normality assumption over the decoder output (Eq. \ref{eq:cVAE_model}). Moreover, the decoder assumes unstructured output noise, with the decoder mean often used as the cVAE output. This treatment implicitly assumes negligible decoder noise. In practice, output noise is often structured \citep{dorta2018structureduncertaintypredictionnetworks} and neglecting it leads to a loss of small‑scale detail \citep{seaicediffFinn2024, Gooya2026}. \\

 \noindent To address this, we employ a framework that makes no parametric assumptions of the decoder's output distribution. As a result, MSE is no longer derived in the objective function (Eq. \ref{eq:objective_function}) and the negative log-likelihood based reconstruction term (Eq. \ref{eq:elbo}) does not have a closed from solution. Instead, we reformulate the problem within the Generalized Variational Inference (GVI; \citealp[]{GVI2022}) framework, which extends the choices of loss functions. Specifically, the GVI framework treats inference as an optimization problem  through the Rule of Three (RoT); (1) a generalized loss function that not need to be a negative log-likelihood, (2) a divergence from the prior that not need to be KL, and (3) a set of candidate distributions, or variational family, over which to search for the approximate posterior.  The standard cVAE objective function is the special case with the negative log-likelihood, the KL divergence, and a Gaussian family. \citet[Section 4.4.4]{GVI2022} demonstrate that the GVI framework can accommodate variational autoencoders (VAEs) with the joint optimization of variational and generative model parameters. They further emphasize that, in VAEs, the probabilistic encoder primarily serves to regularize the latent representation rather than to approximate a true Bayesian posterior (see Eqs. \ref{eq:log_p(x|y)} and \ref{eq:elbo}). This perspective motivates our replacement of the negative log-likelihood term with another strictly proper scoring rule that similarly targets the entire predictive distribution \citep{Gneiting01032007}, rather than individual statistical properties, while retaining the KL term to regularize the latent representation. The Continuous Ranked Probability Score (CRPS) becomes the natural choice, although this has implications for the ELBO on the log-likelihood of the data-generating distribution, as we discuss at the end of the section. \\

\noindent CRPS is a key metric for evaluating probabilistic forecasts that has gained traction as an objective function for weather and climate applications \citep{alet2025skillfuljointprobabilisticweather, lang2024aifscrpsensembleforecastingusing, Kochkov2024NeuralGCM, zhong2024fuxiensmachinelearningmodel, oskarsson2024probabilisticweatherforecastinghierarchical}. Unlike MSE that requires a single output (i.e., the mean of the Gaussian distribution), calculating CRPS for a non-parametric distribution (Eq. \ref{eq:crps}) requires an \textit{output} ensemble $\hat{Y}^{M}_{tl}$ 
to be compared with the target observation ${y}_{tl}$. Previous studies using CRPS for training \citep{Koochali2025, zhong2024fuxiensmachinelearningmodel} or fine-tuning \citep{oskarsson2024probabilisticweatherforecastinghierarchical} VAE models acquire this output ensemble \textit{during training} by decoding, with the decoder in Eq. \ref{eq:cVAE_model}, the $M$ different samples from the posterior distribution $q_{\phi}(\cdot)$, or the prior distribution $p_{\omega}(\cdot)$ in the fine-tuning stage. In such a context, the CRPS acts on the latent space to calibrate the predictions. However, these studies do not explain the connection to the mathematical formulation of cVAE as a Variational Inference problem. Our work is different in that we acquire the output ensemble for \textit{each} ${z}^{n}$ from the latent space and formulate the objective function under the GVI framework. We are not aware of another such a study for climate applications.\\

\noindent In our formulation, we replace the Gaussian decoder $p_{\theta}({y} | z, \bar{x}_{tl})$ in Eq. \ref{eq:cVAE_model} with a probabilistic generator $G_{\theta}(z, \bar{x}_{tl}, \epsilon)$. This allows us to generate an ensemble of outputs for each generated sample ${z}^{{n}}_{tl}$ in the latent space, needed to calculate the CRPS. The generator is therefore optimized to output samples from a (non-parameterized) distribution $F_\theta(y | z, \bar{x}_{tl})$ rather than the parameters of a Gaussian (i.e., mean and standard deviation) as in standard cVAEs. Specifically, instead of a Gaussian output distribution, we employ samples $G_{\theta}(z^{n}_{tl}, \bar{x}_{tl}, \epsilon) \sim F_\theta(y | {z}^{n}_{tl}, \bar{x}_{tl})$. The  objective function then reads (compare with Eq. \ref{eq:objective_function}):

\begin{equation}
    \mathcal{L}(x;\theta,\phi) = \beta\mathrm{KL} \Bigl(q_{\phi}(z \mid y_{tl}, \bar{x}_{tl}) \,\Vert\, p_{\omega}(z \mid \bar{x}_{tl}) \Bigr) + \frac{1}{N} \sum_{n=1}^{N} \mathrm{CRPS}(y_{tl},  G^M_{\theta}(z^{n}_{tl}, \bar{x}_{tl}, \epsilon^M )  )
    \label{eq:objective_function_CRPS}
\end{equation}

\noindent with $\beta$ controlling the relative weight of the two terms and $G^M_{\theta}(z^{n}_{tl}, \overline{x}_{tl}, \epsilon^M)$ indicating the $M$-ensemble of outputs generated for each sample ${z}^{n}_{tl}$ using $M$ independent noise injections ($\epsilon^M$) as explained below. \\

\noindent Drawing on the conditional-GAN framework \citep{daust2024capturing}, we transform the cVAE decoder (Eq. \ref{eq:cVAE_model}) into a truly probabilistic generator by using noise-injection layers that induce stochastic behavior (section \ref{sec:methods}\ref{subsec:inference}). During training, we generate 10 outputs for each latent sample ${z}^{n}_{tl}$. CRPS is then calculated at each grid cell using Eq. \ref{eq:crps} for the 10-member ensemble and is averaged on the spatial dimension. In a standard cVAE with spherical Gaussian decoder (Eq. \ref{eq:cVAE_model}), the log-likelihood over the joint spatial distribution factorizes into summation over grid-cells. For CRPS, however, the spatial avenging amounts to grid-level only (marginal-only) optimization \citep{alet2025skillfuljointprobabilisticweather}. We refer to this model as cVAE-CRPS. The decoder in cVAE versus cVAE-CRPS can be regarded as mapping ${z}^{n}_{tl}$ samples from the lower dimensional latent space to a Gaussian distribution (cVAE) versus a collection of generated points (cVAE-CRPS) in the higher dimensional output space. The cVAE-CRPS model is therefore formulated as follows:

\begin{equation}
    \begin{aligned}
        &\textit{Encoder:} \quad q_{\phi}({z} | {y_{tl}}, {\bar x_{tl}}) = \mathcal{N} \left( {\mu}_{NN_{\phi}}({y_{tl}}, {\bar x_{tl}}), {\sigma}_{NN_{\phi}}^2({y_{tl}}, {\bar x_{tl}}) {I} \right) \\
        &\textit{Decoder:} \quad F_\theta(y | z, \bar{x}_{tl}) \sim G^M_{\theta}(z, \overline{x}_{tl}, \epsilon) \\
        &\textit{Prior:} \qquad p_{\omega}({z} | {\bar x_{tl}}) = \mathcal{N} \left( {\mu}_{NN_{\omega}}({\bar x_{tl}}), {\sigma}_{NN_{\omega}}^2( {\bar x_{tl}}) {I} \right)\\
    \end{aligned}
    \label{eq:cVAE_CRPS_model}
\end{equation}

\noindent A caveat of formulating cVAE-CRPS under GVI is that its objective function (Eq. \ref{eq:objective_function_CRPS}), unlike the standard cVAE objective (Eq. \ref{eq:elbo}), is not a lower bound on the log-likelihood of the generative distribution. 
In the standard cVAE, optimization reduces the gap between the approximate and true posteriors (see Eq. \ref{eq:log_p(x|y)}), while the KL term pulls the aggregate posterior (i.e., the posterior averaged over the population data) toward the prior (see Eq. \ref{eq:elbo}) \citep{hoffman2016elbo} with both identities holding under ELBO tightness (Sec. 2a) at population-level optimum \citep{yacoby2020characterizing}. Together, these align training with generation with the data generating distribution, i.e., sampling $z$ from the prior and decoding yields data approximately consistent with the data-generating distribution. Under cVAE-CRPS, the ELBO interpretation and tightness guarantee are lost. However, the KL term continues to encourage agreement between the latent distributions encountered during training and inference.  Consequently, if the aggregate posterior matches the prior, an argument analogous to that of the standard cVAE can be made at the population level: sampling $z$ from the prior is then equivalent to sampling from the aggregate posterior, and the strict propriety of CRPS drives the generator toward the true conditional distribution of the data given $z$, thereby recovering the data-generating distribution at grid-level. In practice, however, such exact matching generally does not hold for either the standard cVAE or our formulation, due to finite training data, finite ensemble size, limited network capacity, and imperfect optimization. Nevertheless, we show empirically that the full-field maps generated by cVAE-CRPS at inference outperform those of the standard cVAE, not only in marginal metrics but also in global structure and physical consistency, even though the model is optimized on marginals.

\subsection{Architecture and inference}
\label{subsec:inference}

\noindent Our cVAE setup follows the architecture of \cite{cVAE2015} with some modifications. The encoder and prior networks are series of double convolution processing blocks using a variation of ConvNeXt blocks \citep{Samudra, ConvNext2020} with partial convolution layers \citep{partialconv} followed by maxpooling downsampling, mapping to a 1000-dimensional latent space. The decoder reverses the encoder and prior network operations using upsampling and double ConvNeXt blocks, followed by a final convolution layer projecting back to the data space. For the cVAE-CRPS model, we inject random noise as a channel in each of the upsampling and processing blocks in the decoder. Following \cite{cVAE2015}, we create an additional deterministic network combining an encoder and a  decoder without a latent space or noise injection. We employ this pretrained network to provide an initial deterministic (average) estimate of the data $\tilde{y}_{tl}$. This initial estimate is added as an extra condition to the prior network, $p({z} | {\bar x_{tl}, \tilde{y}_{tl}}) = \mathcal{N} \left( {\mu}_{NN_{\omega}}({\bar x_{tl},\tilde{y}_{tl}}), {\sigma}_{NN_{\omega}}^2( {\bar x_{tl}},\tilde{y}_{tl}) {I} \right)$, and added to the main decoder output \textit{before} the final convolution layer. This helps the model focus on learning variations around the estimated deterministic average rather than reconstructing the most likely signal. Details of the architecture and training can be found in the appendix \ref{sec:app_cVAE}. At inference time, samples from the prior distribution are passed to the decoder to generate the output ensemble. For every latent sample drawn from the prior, we sample the output distribution once using the stochastic cVAE-CRPS generator effectively producing an output ensemble as large as the number of latent samples. \\

 \noindent The cVAE-CRPS used in this study has 673-million parameters including the pretrained deterministic branch. The model took nearly 3-hours to train on a single NVIDIA A100 GPU before stalled validation loss triggered early stopping. Pretraining the deterministic branch took an extra 40-minutes prior to training the full cVAE-CRPS. Finally, the inference step runs in 1 hour using 32-CPUs to generate 200 ensemble members over the 2019-2023 period.

 \subsection{Statistical benchmark}
\label{sec:data}

\noindent As a benchmark, we employ a lead-month dependent climatological mean correction combined with the member-by-member calibration (MBMC) of \cite{OntheReliabilityandCalibrationofEnsembleForecasts}, including corections to account for the limited ensemble size \citep{Dirkson2026}. Specifically, if we denote the forecasts as $x_{ijml}$ with $i$ indicating the ensemble member, $j$ and $m$ the initial year and month, respectively, and $l$ the lead month, the benchmark corrected forecast reads:

\begin{equation}
x^{*}_{ijml} = \mu_{ml}^{O} + \alpha_{ml} \left( \bar{x}_{jml}-\mu_{ml}^{H} \right)
+ \beta_{ml} \left( x_{ijml}-\bar{x}_{jml} \right)
\end{equation}

\noindent where $\mu_{ml}^{O}$ and $\mu_{ml}^{H}$ denote the lead-month dependent monthly stratified climatological mean of the observational data and ensemble mean hindcast, respectively, over the 30-year period leading to the initial year $j$. Coefficients $\alpha_{ml}$ and $\beta_{ml}$ are chosen so that the calibrated forecasts jointly satisfy the climatological variance equality condition and the spread–error relationship required for well-calibrated forecasts \citep{OntheReliabilityandCalibrationofEnsembleForecasts} for general ensemble sizes following \cite{Dirkson2026} but for monthly stratified seasonal statistics according to Eq. (\ref{eq:mbmc_alpha}) and (\ref{eq:mbmc_beta}). \\

\noindent This transformation was chosen as a general calibration method up to second order \citep{Dirkson2026} that preserve the ensemble members, similar to cVAE-CRPS. Although more specialized methods exist for SIC forecast post-processing, such as those of \cite{CalibrationSubseasonalSICForecasts, CalibratedProbabilisticForecastsofArcticSeaIceConcentration}, our goal is not to prove the best performing Arctic SIC correction model, but to introduce cVAE-CRPS as a general method for adjustment of seasonal predictions with Arctic SIC as a case study for comparison with, and improvement upon GSA2025.

\section{Data}
\label{sec:data}

\noindent We post-process retrospective seasonal forecasts (or hindcasts) of monthly mean Arctic SIC produced with CanSIPSv3's CanESM5 model. We use raw 10-member ensemble forecasts of 12-month predictions initialized at the beginning of every month starting in January 1981 until the present. The ensemble forecasts are provided on a $1 \times 1$ standard resolution. To train the cVAE-based model, we use hindcasts spanning January 1980 to December 2015 and reserve the forecasts issued in January $2016$ to December $2018$ as a validation set. Adjusted forecasts are tested for years $2019$ to $2023$. We employ a temporal mask to ensure that there is no data leaking from the test sample to the validation sample, or from the validation to the training sample \citep{Sospedra_Alfonso_2024}. \\

\noindent The target observational data are from the satellite-based NOAA/NSIDC Climate Data Record of passive microwave SIC v5 \citep{NOAA-NSIDC-v5} spanning January 1981 to December 2024. These data are provided on the NSIDC Sea Ice Polar Stereographic North grid at 25 km resolution. We re-arrange the observational data into a structure analogous to the forecasts consisting of monthly values spanning 12 months from the start of every month. In addition, we project the forecasts on the same polar stereographic grid as the observations. This implies that for latitudes south to ~$80^\circ$ where the sea-ice edge lies in most months, the cVAE model learns the fine scale variability conditioned on a coarser resolution field. \\

\section{Evaluation metrics}
\label{sec:evaluation_metrics}

\noindent The corrected ensemble using cVAE-CRPS (hereafter Nadj) is compared to the benchmark ensemble (hereafter Badj) and the GSA2025 standard cVAE (hereafter Nadj\_mse). The Badj ensemble is limited by the number of members in the raw ensemble forecasts (10 in this case). For consistency, we evaluate the Nadj and Nadj\_mse ensembles using 10 members only, but emphasize that its size can be made arbitrarily large. For an estimation of uncertainty in the skill metrics, we repeat each calculation 10 times using randomly sampled ensembles for each method. For Nadj and Nadj\_mse, this means generating 10 separate 10-member ensembles using the prior. For Badj, the 10 ensemble members are sampled randomly with replacement 10 times. Moreover, the target satellite-based NOAA/NSIDC SIC data have 2-decimal point numerical precision and masks SIC values below 0.1. We apply the same conversions to the output of all adjusted products. \\

\noindent We assess the prediction skill of marginal (grid-cell-wise) distributions using the RMSE of the ensemble mean forecast and CRPS (Eq. \ref{eq:crps}) relative to the observational data (hereafter Obs), averaged spatially over valid grid cells and over initialization times. Grid cells are not valid if all forecast ensemble members and observations are simultaneously 0 or 1. This mask is applied when comparing to the observation to avoid biased results toward open ocean or fully ice covered regions, respectively. We employ rank histograms (RH) to assess whether a necessary (although not sufficient) condition for a well-calibrated forecast is satisfied \citep{Dirkson_2025}, meaning that  observations and ensemble members are statistically indistinguishable.. Specifically, we count the rank of the Obs relative to the sorted ensemble forecast members at each valid grid point, initialization time and lead month, group the ranks to create a histogram, and report the Cumulative Distribution Function (CDF) \citep{daust2024capturing}. For a calibrated forecast, each rank should have the same probability of occurrence, so the histogram should be a uniform distribution and its CDF the 1:1 straight line. If the CDF has more weight at the tails (peak), corresponding to a U-shaped (reversed U-shape) histogram, the ensemble forecast is overconfident (underconfident). We emphasize that uniform RH is a necessary but not sufficient condition for a statistically calibrated forecast as an unreliable forecasts might satisfy uniform RH due to departures from reliability balancing over the full set despite existence of climatological biases \citep{Dirkson2026}. To complement RH, we compare the climatological distribution of sea ice concentration values pooling all valid grid cells and initialization times using quantile-quantile (QQ) plots.\\

\noindent Additionally, we compare the Spread over Error (SOE) ratio, as a measure of reliability of the ensemble forecasts up to second order. SOE is defined in terms of the ensemble variance and the MSE of the ensemble mean forecast (Eq. \ref{eq:soe}) after removing the monthly climatology and thus effectively removing forecast climatological bias. As above, SOE=1 is a necessary condition for a well-calibrated forecast \citep{SOE2013}, but not  sufficient since SOE $=$ 1  must be accompanied by an unbiased climatological mean, variance, and inter-member covariance for a perfect reliability \citep{Dirkson2026}. Therefore, we include computations of those quantities along with SOE. We compute the lead-month dependent climatological mean bias as the difference between forecast and observed climatology. The lead-month dependent climatological variance bias, denoted as $\Delta\sigma^2$, is computed as the difference between the variance in observations and the ensemble total variance, the latter defined as the ensemble mean over the variance of each ensemble member over the analysis period. Similarly, the inter-member covariance bias, denoted $\Delta\Sigma$, is computed as the difference between the average pairwise member covariance and ensemble-averaged covariance of each ensemble member and observations. When reporting spatial averages, we compute absolute values first to avoid bias cancellation through summation across the spatial domain. We compute the SOE ratio and individual bias terms at each valid grid point and average spatially similar to RMSE and CRPS. Details are provided in appendix \ref{sec:app_metrics}. \\

\noindent Finally, we evaluate the corrected ensemble using several metrics that reflect the spatial and temporal structure of the forecasts. We compute RMSE and anomaly correlation coefficient (ACC) of sea ice area (SIA, Eq. \ref{eq:SIA}) and extent (SIE, Eq. \ref{eq:SIE}), as well as pattern correlation averaged over the initialization times. We compute the Integrated Ice Edge Error (IIEE, Eq. \ref{eq:IIEE}), which measures the difference between the areas enclosed by predicted and true ice edges probabilities \citep{IIEE}. Furthermore, we use the radially averaged power spectral density (RAPSD) to compare the covariance structure of the adjusted forecasts and the verification data, and to diagnose the smoothness of cVAE outputs \citep{alet2025skillfuljointprobabilisticweather}. For each target month, RAPSD is calculated by annularly averaging the two-dimensional Fourier spectra around the origin of the wavenumber domain.

\section{Results}
\label{sec:results}

\begin{figure}
\centering
\includegraphics[width=\textwidth]{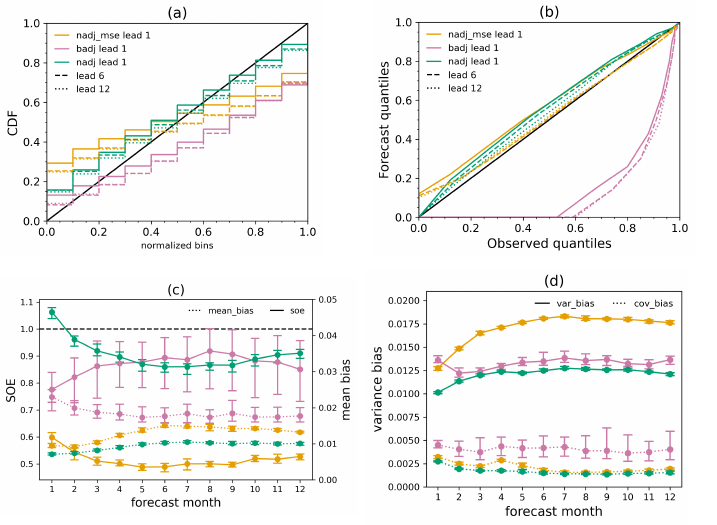}
\caption{Comparison of climatological distribution of 10-member Nadj (green), Nadj\_mse (yellow) and Badj (magenta) ensembles against Obs over the test period 2019-2023 as a function of forecast month. (a) CDF of RH and (b) QQ plots for lead months 1, 6 and 12. Computations are performed over valid grid  cells as defined in section \ref{sec:evaluation_metrics}. (c) SOE (left y-axis, solid curves), climatological mean bias (right y-axis, dotted curves), and (d) climatological bias in variance (solid curves) and inter-member covariance (dotted curves) as a function of forecast month computed from anomalies relative to climatology. Error bars indicate minimum and maximum values based on randomly sampling 10 ensemble members with replacement.}
\label{fig:fig1}
\end{figure}

\noindent Figure \ref{fig:fig1} shows properties of the climatological distribution of the 10-member adjusted forecasts relative to the observations over the test period. For the Nadj ensemble, the CDF curves at various lead months are closer to the 1:1 line than the benchmarks (Fig. \ref{fig:fig1}a), corresponding to RHs closer to the uniform distribution for calibrated ensembles. The Nadj\_mse ensemble shows a smaller bias than Badj (symmetry relative to the 1:1 line), although its smaller slope indicates an U-shaped RH suggesting an overconfident ensemble. Badj, on the other hand, has largely asymmetric CDFs indicative of biased low ensembles, with observations more frequently lying in the higher ranks. This is confirmed with QQ plots showing Badj consistently underestimating the observed quintiles (Fig. \ref{fig:fig1}b). Zero-inflated quintiles imply that Badj ensembles predict ice-free conditions regularly where ice is observed, indicative of errors in predicting the edge of the sea ice cover. On the other hand, Nadj\_mse QQ curves show that the lower observational quintiles are overestimated, indicative of predicted ice in regions of little or not observed sea ice. Nadj consistently shows a better match with observation and thus a closer climatological distribution.\\

\noindent SOE computed with anomalies relative to climatology is less than one for Nadj and Badj for most forecast months, with Badj having a greater uncertainty (Fig. \ref{fig:fig1}c). For Nadj, values relatively close to one are seen for the first forecast months, whereas Badj has values closer to one later in the forecast. As discussed in section \ref{sec:evaluation_metrics}, this doesn't guarantee reliability, so we further examine the climatological bias in the variance, $\Delta\sigma^2$, and inter-member covariance, $\Delta\Sigma$. Figure \ref{fig:fig1}d shows that $\Delta\sigma^2$ and $\Delta\Sigma$ are consistently smaller for Nadj than the benchmarks, by a relatively large margin for $\Delta\Sigma$, suggesting that SOE $ \approx 1$ for Badj might not result from a more reliable forecast. For example, forecasts with relatively high biases but satisfying $\Delta \sigma^2 = 2\Delta\Sigma$ may have SOE $ = 1$ without being reliable \citep{Dirkson2026}. Fig. \ref{fig:fig1}c furthermore shows smallest mean bias in Nadj for all forecast months. We note, however, that our estimates of climatological mean and consequently the anomalies used to compute SOE are based on a relatively small 5-year sample, and can therefore be subject to considerable sampling error. Ideally, the test period would be large enough to have a 30-year base climatology for an out-of-sample evaluation, which is the standard for seasonal forecasts. Alternatively, a leave-one-out approach could be used, as documented in standard calibration studies \citep{Kharin2017}, but this would come at a high computational cost (i.e., one cVAE model must be trained for each leaved-out year) and it is beyond the scope of this study. Based on the 5-year estimates, Nadj is closer (and not worst) in climatological distribution to the target than Badj up to second order despite their SOE being statistically similar and smaller than one. Finally, the large variance bias in Nadj\_mse results in lowest SOE despite having smaller mean bias and $\Delta\Sigma$ than Badj, consistent with overconfident RH and a muted variability typical of standard cVAE, as we discuss below. \\

\begin{figure}
\centering
\includegraphics[width=\textwidth]{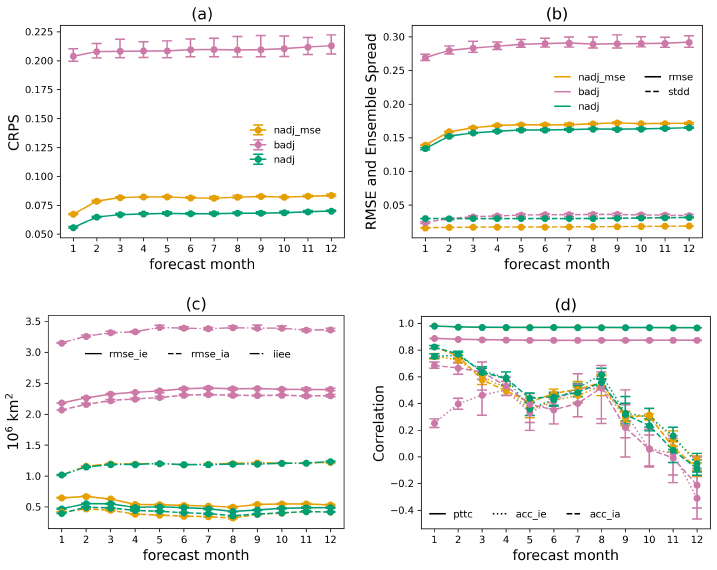}
\caption{Comparison of 10-member Nadj (green), Nadj\_mse (yellow) and Badj (magenta) individual ensemble forecasts against Obs over the test period 2019-2023 as a function of forecast month. Spatially averaged (a) CRPS and (b) RMSE (solid curve) over valid grid cells as defined in section \ref{sec:evaluation_metrics}. The dashed curve in (b) shows ensemble standard deviation computed first at grid-cell level, then averaged over the entire region. (c) RMSE of SIA (solid curve) and SIE (dashed curve), average IIEE (dotted curve), (d) ACC of  SIA (dashed curve) and SIE (dotted curve), and pattern correlation (solid curve) as a function of forecast month. ACC is computed along initial times and all other quantities are averaged over initial times in the test period. Error bars indicate minimum and maximum values based on randomly sampling 10 ensemble members with replacement.}
\label{fig:fig2}
\end{figure}

\noindent Figure \ref{fig:fig2} compares individual ensemble forecasts (at each initial-time and forecast month) against the observation for the different adjustment methods using several measures of performance as a function of forecast month. Overall, Nadj tends to outperform the benchmark for each measured consider. In terms of CRPS (Fig. \ref{fig:fig2}a), Nadj is better calibrated and sharper than Nadj\_mse, and particularly Badj. This is consistent with the zero-inflated QQ plots shown in Fig. \ref{fig:fig1}b and the deterministic measures shown in Fig. \ref{fig:fig2}b,c. Indeed, the ensemble mean forecasts of Nadj and Nadj\_mse outperform Badj both at the grid cell level and across the sea ice domain. Figure \ref{fig:fig2}b shows the RMSE of ensemble mean averaged over valid grid cells as a function of the forecast months, indicating smaller errors for both cVAE-based forecasts than for Badj. The average ensemble standard deviation (indicative of noise variance) is also shown on the same panel for each method, computed over non-land grid cells. Interestingly, the ensemble spread for Nadj\_mse is smaller than for Nadj and Badj, which have a similar spread. Actually, Nadj\_mse ensemble spread is muted even for 200 members (not shown), consistent with the documented suppressed fine scale variability and characteristic blurry outputs in standard cVAE models \citep{Gooya2026}. This motivated the post-hoc calibration in GSA2025 through scaling the prior's standard deviation by sampling a wider range of internal variability \citep{gooya2025probabilisticbiasadjustmentseasonal}, and led to this study on the cVAE-CRPS model. As we show below, the cVAE-CRPS design results in a richer and sharper ensemble profile as well as sharper output maps.\\

\noindent For the bulk measures of integrated sea ice, Fig. \ref{fig:fig2}c shows the RMSE of the sea ice area (SIA, Eq. \ref{eq:SIA}) and extent (SIE, Eq. \ref{eq:SIE}) as well as the time-averaged errors over the ice edge (IIEE, Eq. \ref{eq:IIEE}) reflecting global structure. For all metrics, the ensemble mean Nadj and Nadj\_mse forecasts outperform Badj by a large margin. Specially for IIEE, the poor performance of Badj at predicting sea ice edge is consistent with its distorted QQ plot structure discussed above (Fig. \ref{fig:fig1}b). Similarly, the pattern correlation averaged over time is improved for the ensemble mean Nadj and Nadj\_mse forecast relative to Badj (Fig. \ref{fig:fig2}d). Despite being optimized at the marginal grid-cell level, these results highlight the benefits of cVAE methods to capture the joint spatial structure and modeling fully resolved spatial maps compared to independent grid-cell corrections. In the case of ACC, on the other hand, all adjusted forecast ensembles have statistically similar performance, with Badj showing worse skill for IE early in the forecast. Similar to SOE and its decomposition, ACC results are highly susceptible to sampling error as the test period is only five years ($2019-2023$). We note that the cVAE is not explicitly tasked with learning the forecast time dependence. The phase of the interannual variability in both the Nadj and Nadj\_mse forecasts is learned through the predictable component in the condition, given by the ensemble mean of the (biased) raw forecast. Therefore, there is no expectation that Nadj and Nadj\_mse would significantly outperform Badj in terms of ACC unless it is provided with extra information (e.g., from other variables or conditioning fields). Remarkably, the cVAE model largely captures the time dependence of the integrated quantities while generating more accurate and reliable ensemble forecasts despite its conditioning field being biased. \\

\noindent We further assess the spatial covariance and physical realism of the generated SIC fields using the RAPSD metric (Fig. \ref{fig:fig2}d,e). Specifically, we compute the RAPSD ratios of the adjusted forecasts relative to the Obs for the \textit{target} months of March and September (Fig. \ref{fig:fig2}d), representative of the monthly maximum and minimum Arctic sea ice cover, respectively, for the test period $2019-2023$ and using 10-member ensembles for all adjusted forecasts. On average across all initial times and ensemble members, Badj exhibits spectral-energy biases at both large and small spatial scales for the two target months and lead times considered. The spectral errors at large scales are likely consequence of the artifacts introduced by grid-cell adjustment that neglects the spatial covariance structure, whereas the loss of fine-scale energy reflects the coarse resolution of the raw forecasts relative to the target observations. By contrast, Nadj exhibits relatively small spectral-energy biases at both large and small spatial scales. This is noteworthy, as the post-processing tool not only corrects mean-state biases and calibrates the raw forecast ensemble but also effectively increases the spatial resolution by leveraging the finer resolution of the target observations. The increased spatial resolution is evident in Fig. \ref{fig:figA1}, which shows a randomly chosen ensemble member for Nadj and Badj forecasts, as well as the target observation as a reference, for September and March $2023$.\\

\begin{figure}
\centering
\includegraphics[width=\textwidth]{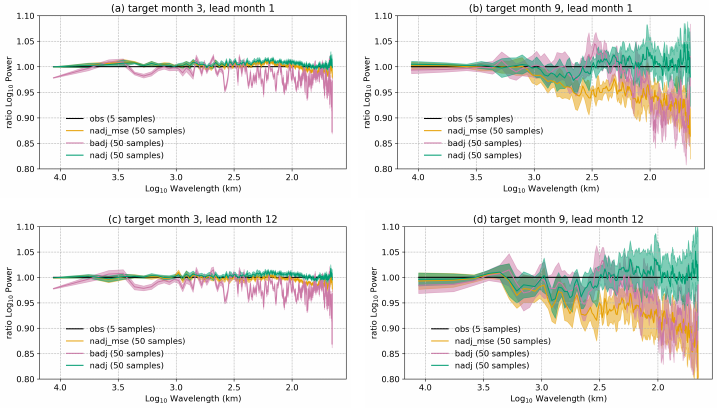}
\caption{ RAPSD ratios versus Obs for Nadj\_mse, Badj and Nadj for predictions of (a) March maximum on lead month 1 averaged over $2019 - 2023$ (5 years) and across ensemble members (5 $\times$ 10 samples). The shading shows the range associated with different members. (b) same as (a) but for September minimum. (c) and (d) same as (a) and (b) but for lead month 12.
} 
\label{fig:fig3}
\end{figure}

\noindent One key motivation to modify the original cVAE model of GSA2025 is its muted variability and the spectral-energy bias exhibited at fine scales. The RAPSD curves for Nadj\_mse show that cVAE-CRPS does a better job at representing fine-scale variability  (Fig.  \ref{fig:fig3}), while standard cVAE has a clear drop in energy at fine scales. The improvements are seen for both March and September sea ice, as well as for short and long lead months. The loss of fine scale variability and smoothness of outputs becomes evident in SIC maps (Fig.  \ref{fig:figA1}) which is greatly improved with cVAE-CRPS. This is while cVAE-CRPS outperforms cVAE\_mse for nearly all deterministic and probabilistic metrics considered in this study. \\

\noindent At fine spatial scales, all adjusted forecasts underperform for the September sea ice minimum and present large uncertainties (Fig. \ref{fig:fig3}b,d). The Badj and Nadj\_mse ensembles show larger drops in RAPSD ratios compared to March maximum at scales smaller than 100 km. By contrast, Nadj ratios remain closer to 1 but with a tendency to overshoot fine-scale variability. On longer lead months, RAPSD ratios show consistent performance on average, but with larger uncertainty reflected in the wider range values across the ensemble members (see shading in Figs. A6 and A3). This is consistent with earlier results showing higher uncertainty for predictions over longer lead months. For SIC, which is bounded between 0 and 1, the uncertainty in RAPSD ratios likely reflects uncertainty in representing the transitional regions between fully ice-covered and open-water conditions, particularly near the marginal ice zone and ice edge. In this regard, the RAPSD inaccuracies and increased uncertainty during the September minimum would be in agreement with previous studies showing reduced predictability of Arctic sea ice concentrations during summer and increased variability of the ice edge during the melt season \citep{sigmond2016}. This is shown below with geographic maps of the ensembles spread. The overshoot in variability at fine scales during the September minimum for cVAE-CRPS may be related, at least in part, to the noise injection in the last up-sampling layer (Appendix A1), which may add to an already noisy forecast during the melt season, particularly at the sea ice edges. It shouldn't be ruled out, however, that despite having a demonstrated success in capturing spatial covariance structure, CRPS optimization is done at marginal grid-cell level, which may also contribute to the excess in variability similarly documented in other studies \citep{alet2025skillfuljointprobabilisticweather}.\\

\noindent Figure \ref{fig:fig4} shows the maps for the ensemble standard deviations for the target months of September and March, averaged over the test sample. This variability reflects the underlying forecast uncertainty and is expected to be greatest along the sea-ice edge, where freezing and melting processes tend to fluctuate more strongly. The figure shows that both Nadj and Nadj\_mse ensemble forecasts capture this behavior, with the variance being stronger along the observed sea ice edges (dotted lines in the figure indicating $SIC \geq 0.15$). Unlike Nadj\_mse, which presents a muted variability, Nadj variability is consistent with Badj in magnitude while exhibiting sharper spatial coverage. The spread grows with lead months, both in magnitude and spatial coverage (Fig. \ref{fig:fig4}), consistent with greater uncertainty in their RAPSD ratios (Fig. \ref{fig:fig3}c,d), especially at small spatial scales, and the general tendency for forecast noise variance to increase with lead time. For Badj, the uncertainty in September sea ice at a 12-month lead is particularly striking (Fig. \ref{fig:fig4}), as it encompasses most of the Arctic ocean despite the high observed SIC towards the pole indicative of permanent sea ice cover. By contrast, the Nadj uncertainty for the same target and lead month is more closely confined to the observed edges of the sea ice cover (Fig. \ref{fig:fig4}), as expected.\\

\begin{figure}
\centering
\includegraphics[width=\textwidth]{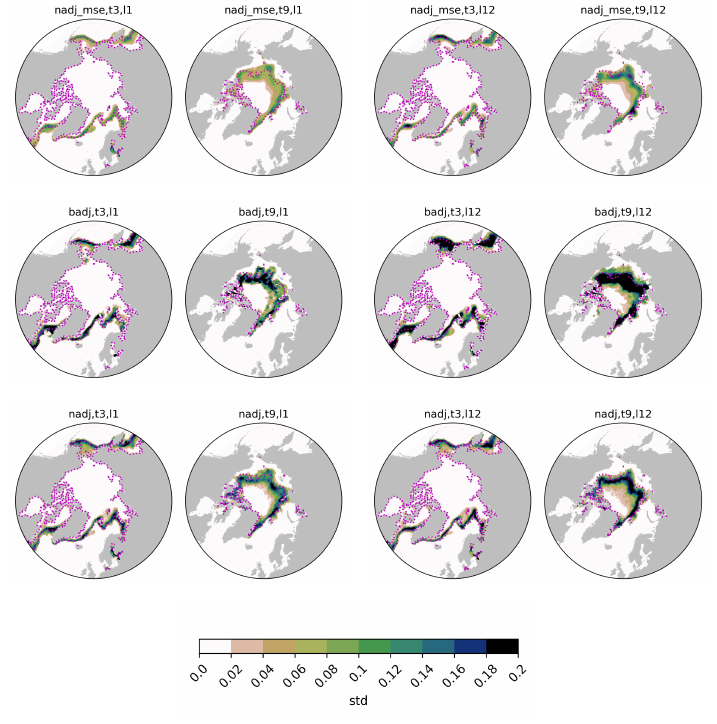}
\caption{Standard deviation (std) across ensemble averaged for predictions on target months of March maximum (first and third columns, t3), and September minimum (second and fourth columns, t9) on lead months 1 (l1 in first and second columns), and 12 (l12 in third and forth columns)  over $2019-2023$. The rows show different product being Nadj\_mse, Badj and Nadj, respectively. The dashed magenta line shows the mean ice edge from observations over the same period. } 
\label{fig:fig4}
\end{figure}

\section{Discussions and conclusions}
\label{sec:discussions_and_conclusions}

\noindent This work presents a probabilistic post-processing framework developed at the Canadian Centre for Climate Modeling and Analysis (CCCma) to generate arbitrarily large ensembles of adjusted, high-resolution seasonal forecasts. The proposed methodology builds on the conditional Variational Autoencoder model of \cite{gooya2025probabilisticbiasadjustmentseasonal} to adjust seasonal predictions of Arctic Sea Ice. The key innovation in this work is replacing the cVAE's Gaussian‑parametrized decoder with a generator that learns a non-parametric output distribution. Training is carried out using CRPS as a a Generalized variational Inference problem which is different than the original ELBO formulation of standard cVAEs. By moving away from the Gaussian assumption and thus from MSE-based objective function, this approach preserves fine-scale structure and avoids the overly smooth outputs that typically affect conventional cVAEs. \\

\noindent We apply the cVAE-CRPS model for the probabilistic adjustment of seasonal predictions of Arctic Sea Ice Concentrations produced with CCCma's CanESM5 model. We used several deterministic and probabilistic metrics to evaluate the performance for marginal distributions (grid-cell-level) and to capture the spatial structure of the adjusted forecasts relative to a generalized climatological mean corrected MBMC adjustment as well the standard cVAE as benchmarks. The forecast ensemble adjusted with cVAE-CRPS consistently outperforms the benchmarks for all metrics and lead months by a considerable margin. Specifically, the cVAE-CRPS adjusted ensemble produces RH that are nearly uniform (Fig. 1a), and have reduced climatological mean, variance and inter-member covariance biases than those of the benchmark adjustments (Fig. \ref{fig:fig1}d). It shows strong agreement with the verification data in the distribution of SIC evidenced with QQ plots (Fig. 1c), particularly over the ice edges where uncertainty is greatest, and it shows substantial CRPS improvements across the ensemble (Fig. 1d). In addition, the model significantly reduces errors in the ensemble mean forecast both at grid cell level (Fig. 2) and for integrated measures of sea ice area and extent across the Arctic domain, notably in predicting the edges of the sea ice cover. Importantly, the cVAE-CRPS model reduces spectral-energy bias not only at large spatial scales but also at fine scales (Fig. 2d,e), successfully reconstructing the target's fine-scale variability which the benchmark fails to achieve. The tendency for the cVAE-CRPS spectral energy to overshoot at fine scales in melt season results from the combination of feature-level noise injection and CRPS optimization for marginal (grid-level) distribution. The former can potentially be remedied by injecting noise in fewer layers after the final upsampling block (i.e., the finest spatial scale; see Appendix A1) in the generator. Alternatively, noise can be injected in parameter space instead of feature channels \citep{alet2025skillfuljointprobabilisticweather}. In case of the latter, using multivariate proper scoring rules such as the energy score, a generalization of the CRPS to multivariate distributions, could directly target the joint spatial distribution and thereby directly optimize for spatial dependence across grid cells \citep{Gneiting2008}.\\

\noindent The cVAE-based approach formulates the probabilistic adjustment of seasonal forecast ensembles by modeling the target observational distribution conditioned on the biased ensemble mean forecast from the dynamical model, in this case CCCma's CanESM5. The ensemble mean forecast carries information of CanESM5's predictable component due to both external forcing and internal variability, although it drifts toward the model preferred climatology typically reducing its predictive capacity. The cVAE thus models the distribution of realizable outcomes in the observational distribution given this (biased) signal from the climate model. Based on preliminary analysis of the latent space (not shown), we hypothesize that the cVAE performs adjustment by identifying \textit{analogue} representations in the observations, via the conditional prior distribution, given the biased ensemble mean forecast. However, unlike traditional ensemble analogue methods that search the observational record directly in data space, the cVAE has the potential to identify analogues within a structured lower-dimensional latent space. This offers a key advantage: in the structured and continuous latent space of a cVAE, the network learns to more effectively disentangle the condition's relevant features, enabling it to infer analogues from a continuum of lower-dimensional, observationally-based structured representations. Crucially, the prediction skill of cVAE-CRPS is limited by the conditionining field, as evident with ACC results, unless additional information is provided, e.g., from other input variables or conditions. \\

\noindent The primary advantage of cVAE models and generative machine learning approaches more broadly over traditional post-processing methods is that they are fundamentally probabilistic methods that model the high dimensional joint spatial distribution of the data. This is especially important for chaotic systems such as weather and climate that are inherently uncertain. A proper representation of the full range of possible outcomes in the high-dimensional output space is thus essential for informed decision making. This is where possible challenges with underdispersivity in generative models merit special attention. Variability in cVAE based models comes potentially from two sources; stochasticity in latent space sampling constrained by the prior distribution, and uncertainty in the output space formulated in the decoder distribution. For the former, GSA2025 showed that the explicit mathematical formulation of cVAEs, combined with their continuous and structured latent space, enables controlled sampling over a broader range of internal variability through scaling the prior's standard deviation. However, caution should be taken not to overcompensate the muted fine scale output noise by wider range of internal variability through latent space as the two have different spatial scales. For the output noise, \cite{Gooya2026} showed that ignoring the variability in the decoder output, e.g., by taking the mean of a Gaussian decoder as in traditional cVAEs, leads to smooth fields where fine scale structure is lost (section 5). To address this, we implemented a cVAE-CRPS framework that enables the generation of sharp SIC maps from the output distribution rather than assuming an explicit Gaussian form. As a consequence, we were able to maintain fine-scale energy and produce sharp outputs and richer ensemble spared without scaling the prior distribution. Furthermore, this model is naturally suitable for downscaling, which is another important application for reliable high-resolution forecasting. \\

\noindent Our results show that the cVAE-CRPS model is able to generate more skillful, reliable, and calibrated forecast ensembles than the climatological MBMC correction and the standard cVAE model of GSA2025. The results are based on a 10-member cVAE-CRPS ensemble generated to mirror the benchmark ensemble, ensuring a fair comparison. However, the size of the cVAE-CRPS ensemble output can be increased arbitrarily, which is possible in the framework of generative, probabilistic post-processing methods. Importantly, this framework is flexible enough to allow conditioning on several variables or to perform a multivariate adjustment. Future work will examine whether incorporating additional variables or covariates, particularly those with higher predictability, can extend the predictability horizon in forecasts of other target variables. A leave-one-out approach might be necessary for acquiring a 30-year test sample for a less noisy estimate of anomalies. Finally, the current framework generates adjusted forecast ensembles without explicit dependence on lead month. In principle, the generated ensembles members should be independent if they sample unpredictable variability. However, due to potential limitations with the optimization process, model architecture, and training data, residual temporal dependencies (e.g., due to higher frequency information than the monthly data) might exist that are not accounted for. To address this limitation, future work will explore sequence models aimed at generating adjusted ensemble forecasts traceable in time.

\acknowledgments

\noindent We thank Mr. Johannes Exenberger for his advice and valuable input during ideation and early stages of this research. The authors declare no conflict of interest.


\datastatement

\noindent The seasonal predictions of \textit{SIC} from CanESM5 using CMIP6 forcing used in this study are not currently publicly available. However, SIC seasonal hindcast data initialized in 1980-2023 used here will be made available at the time of publication in a public repository. Observational verification data from the satellite-based NOAA/NSIDC Climate Data Record of passive microwave SIC v5 are available at \url{https://nsidc.org/data/g02202/versions/5/}. Research grade codes used for training the models can be found at \url{https://github.com/ParsaGooya/SI_seasonal_forecast_adjustment}. A production ready software package is currently being developed and will be ready before publication. All other inquires should be directed to P. Gooya.

\newpage

\appendix

\section{Architecture and Training}
\label{sec:app_cVAE}

\noindent The building blocks of the cVAE model are convolution blocks based on a modified version of ConvNeXt blocks \citep{ConvNext2020}  as used by \cite{Samudra}. We replaced all 2D convolutions with partial convolution \citep{partialconv} layers. This is a natural choice for the Arctic region, where there is an irregular land mask with small islands. The partial convolution layer automatically ignores these regions while processing the data. The encoder and prior networks follow the same architectures. The encoder input ($y_{tl}$, $\bar x_{tl}$) and the prior network input  ($\tilde {y}_{tl}$, $\bar x_{tl}$), where $\tilde { y}_{tl}$ is an initial estimate produced by a deterministic network (details below). In addition, we add three extra conditioning fields, uniformly entering $\sin(\frac{2 \pi}{12} (t + l))$, $\cos(\frac{2 \pi}{12} (t + l))$, and $(\frac{l}{12})$ as input channels where $t$ is the initialization \textit{month at a given year} and $l$ is the lead month. These networks encode their input into the mean (${\mu}_{NN_{\phi}}$ and ${\mu}_{NN_{\omega}}$)  and variance (${\sigma}_{NN_{\phi}}^2$ and ${\sigma}_{NN_{\omega}}^2$) of the latent and prior distributions, respectively. The architecture is as follows:

\[\begin{aligned}
& \bullet \text{Input (5)} 
\rightarrow \text{ $3\times3$ partial convolution (16)} \rightarrow \text{ Layer normalization (16)}
\rightarrow \text{ DoubleConvNeXt (32)}\\& \rightarrow \text{ MaxPool (32)}  \rightarrow \text{ DoubleConvNeXt (64)}\rightarrow \text{ MaxPool (64)}
\rightarrow \text{ DoubleConvNeXt (128)}\rightarrow \\& \text{ MaxPool (128)} 
\rightarrow \text{ DoubleConvNeXt (256)}\rightarrow \text{ MaxPool (256)}
\rightarrow \text{ DoubleConvNeXt (256)}\rightarrow \\& \text{ Layer normalization (256)} \rightarrow \text{ Dense ($2 \times 1000$) }
\end{aligned}
\]

\noindent The decoder takes samples from the latent space and reverses the operations in the encoder (prior) networks using upsampling and double ConvNeXt blocks. The upsampling blocks are composed of bilinear interpolation to double the resolution of the input, followed by a masked convolution with a 3 × 3 kernel smoothing the interpolated fields. The combination of interpolation with convolution results in less checkerboard effects compared to a transposed convolution \citep{Odena2016}. For the cVAE-CRPS's generator, layers of random noise are injected (concatenated) before each convolution operation in the ConvNeXt blocks, as well as in the upsampling blocks after the bilinear interpolation and before the 3 × 3 convolution. Finally, an output block, which is a combination of "layer normalization, ReLu activation, and $1\times1$ convolution", maps the decoder output back to the SIC space. Like in \citep{finn2024diffice}, we use ReLu activation before the last convolution to help improve the representation of continuous-discrete sea-ice processes. The decoder is as follows:

\[\begin{aligned}
& \bullet \text{Latent samples (1000)} \rightarrow \text{ Dense (256)} \rightarrow  
 \text{ Upsampling (256 + Noise)}  \\&\rightarrow \text{ DoubleConvNeXt (128 + Noise)} 
\rightarrow \text{ Upsampling (128 + Noise)} \\& \rightarrow \text{ DoubleConvNeXt  (64 + Noise)} 
\rightarrow \text{ Upsampling (64 + Noise)} \\&\rightarrow \text{ DoubleConvNeXt (32 + Noise)} \rightarrow \text{ Upsampling (32 + Noise)}\\& \rightarrow \text{ DoubleConvNeXt (16 + Noise)} \\&  \rightarrow
\text{ Layer normalization (16)} \rightarrow \text{ ReLu (16) }
\rightarrow  \text{ $1\times 1$ partial convolution (1) }
\end{aligned}
\]

\noindent The deterministic network used to produce the initial estimate of the bias corrected input $ \tilde{y}_{tl}$ feeding into the encoder and prior networks is composed of the same encoder and decoder architectures without latent space (i.e., by removing the last "Layer normalization + Dense" layers in the encoder, and the first "Dense" layer in the decoder) or noise injection. Importantly, the same output block (layer normalization + ReLu + 1$\times$1 convolution) is shared between the generator (decoder) and the deterministic network to encourage realistic outputs from the deterministic model \citep{cVAE2015}. Lastly, the output of the last DoubleConvNeXt layer of the deterministic model (which has 16 channels) is summed to the corresponding layer output in the decoder before passing to the output block. \\

\noindent We first trained the deterministic model separately using MSE as objective function. We found that this pretraining improves the probabilistic skill of the cVAE-CRPS helping it to focus on learning variability around the deterministic average guess. After the pretraining stage, all network were trained end-to-end using Adam optimizer, a cosine decreasing learning rate with a maximum of 0.0001, and an effective batch size of 4 (2 $\times$ 2 gradient accumulation steps). The KL divergence and CRPS (surrogate for log-likelihood) terms in the loss function (Eq. \ref{eq:objective_function_CRPS}) were normalized based on their dimensionalities (1000 for KL and $432\times 304$ for the CRPS). Given that the dimensionality of the output is an $O(100)$ of the latent space's dimension, the KL term was weighed with $\beta = 0.01$ which was annealed linearly from 0 over the first 10 epochs during training \citep{betaannealing}. The loss over the validation set was used as criterion for early stopping with a buffer of 10 epochs. \\

\noindent We found that the spectral energy is sensitive to noise injection in the decoder's upsampling blocks on spatial scales $\lesssim 100$km, especially for predictions in the summer target months. Specifically, excess energy (ratios $>1$ in Fig.\ref{fig:fig2}) is found for September minimum over the validation period when noise is injected at every spatial resolution. We noticed that removing the noise injection layer over the smallest spatial scale (i.e., after the last upsampling block) improves performance. However, to avoid enforcing the spatial scale over which the stochasticity is introduced, we maintained all noise injection layers. Instead, we added a pooled CRPS term to the loss function, i.e., in addition to optimizing CRPS in each grid cell, we calculated a 2 $\times$ 2 average-pooled CRPS and computed the mean of both terms. The CRPS is defined as:

\begin{equation} 
\mathrm{CRPS}( y,\hat{Y}^M)
=
\frac{1}{M} \sum^{M}_{k=1}|\hat{y}_{k} - y| - \frac{1}{2M^2} \sum^M_{k=1}\sum^M_{k'=1}|\hat{y}_k - \hat{y}_{k'}|
\label{eq:crps}
\end{equation}

\noindent where y is the target observation and $\hat{y}_k$ is the $k$-th ensemble member of the cVAE-CRPS $M$-member output ensemble $\hat{Y}^M$. During training, the output ensemble corresponds to $G^M_{\theta}(z^{n}_{tl}, \bar{x}_{tl})$ for each  $z^{n}_{tl}$ sampled from the posterior $q_{\phi}(z|y_{tl}, \bar{x}_{tl})$,  with $M=10$ (Eq. \ref{eq:objective_function_CRPS}). During evaluation, it corresponds to $ G^M_{\theta}(z^n_{tl}, \bar{x}_{tl})$  for each $z^{n}_{tl}$ sampled from the prior $p_{\omega}({z} | {\bar x_{tl}})$, with $M$ arbitrary large. \\

\section{Evaluation Metrics}
\label{sec:app_metrics}

\noindent RH are produced by ranking the verification data (Obs) over the predicted ensemble at each grid cell, sorting the ensemble members in ascending order and ranking the observational target in that sequence. We evaluate the ranks only across valid grid cells masking those where all ensemble members and observation simultaneously predict 0 or 1. This choice is made to avoid results dominated by the many easy-to-predict grid cells with 0 or 1 values, corresponding to open ocean or fully ice-covered regions, respectively. For each lead month, we compute CDFs by pulling the rankings of all grid cells of marginal ice cover and of all initialization months.  \\

\noindent QQ plots comparing quantiles of the adjusted ensembles (Nadj or Badj) with observations, are constructed, for each lead month, from distributions formed by pooling all valid grid cells, all ensemble members, and all initialization months. \\

\noindent SOE ratios for $M$-member adjusted forecast ensembles $\hat{Y}^{M}$ are computed as \citep{SOE2013}:

\begin{equation}
    \begin{aligned}
          \quad  
          \text{SOE} = \sqrt{\frac{M+1}{M} \frac{\sigma^2_{\hat{Y}^M}}{\hbox{MSE}}}\quad
    \end{aligned}
    \label{eq:soe}
\end{equation}

\noindent where MSE denotes the mean square error of the ensemble mean and $\sigma^2_{\hat{Y}^M}$ the variance of the ensemble around the ensemble mean anomalies relative to monthly climatology, averaged over initialization times. This calculation is done at each valid grid cell and the area weighted average is reported at each lead month. RMSE (over valid grid cells) and standard deviation are calculated as square root of MSE and square root of variance above before averaging, both of which are reported as spatial averages. CRPS is calculated using Eq. \ref{eq:crps} and reported following the same averaging protocol as RMSE over valid grid cells. \\

\noindent For area integrated measures, the total sea ice area (SIA) is: 

\begin{equation}
    \begin{aligned}
          \quad  
          \text{SIA}_{tl} = \int_{Arctic} \text{SIC}_{tl} \quad da\quad
    \end{aligned}
    \label{eq:SIA}
\end{equation}

\noindent where SIC denotes sea ice cover for observations $y_{tl}$ or the ensemble mean $\bar {\hat y}_{tl}$ of Nadj or Badj. 

Similarly, sea ice extent (SIE) is computed as an area integral of grid cells with SIC$>0.15$:

\begin{equation}
    \begin{aligned}
          \quad  
          \text{SIE}_{tl} = \int_{Arctic} I (\text{SIC}_{tl}>0.15) \quad da\quad
    \end{aligned}
    \label{eq:SIE}
\end{equation}

\noindent where $I(.)$ denotes the indicator function. \\

\noindent IIEE, which captures the differences along the ice edges, quantifies the area where the predicted and true ice concentrations differ \citep{IIEE}:

\begin{equation}
    \begin{aligned}
          \quad  
          \text{IIEE}_{tl} = \int_{S_{\geq 50^{\circ}N}} |I (y_{tl}>0.15) - I (\bar{\hat y}_{tl}>0.15)| \quad da\quad
    \end{aligned}
    \label{eq:IIEE}
\end{equation}

\noindent SIA, SIE and IIEE, as well as pattern correlations, are all calculated at each lead month and initialization time, then averaged over initialization times. Finally, for both the SIE and SIA estimates, anomalies relative to monthly climatology are used to calculate the ACC, defined for each lead month as the Pearson correlation coefficient across the initialization time dimension.\\

\begin{figure}
\renewcommand{\thefigure}{A\arabic{figure}}
\setcounter{figure}{0}
\centering
\includegraphics[width=\textwidth]{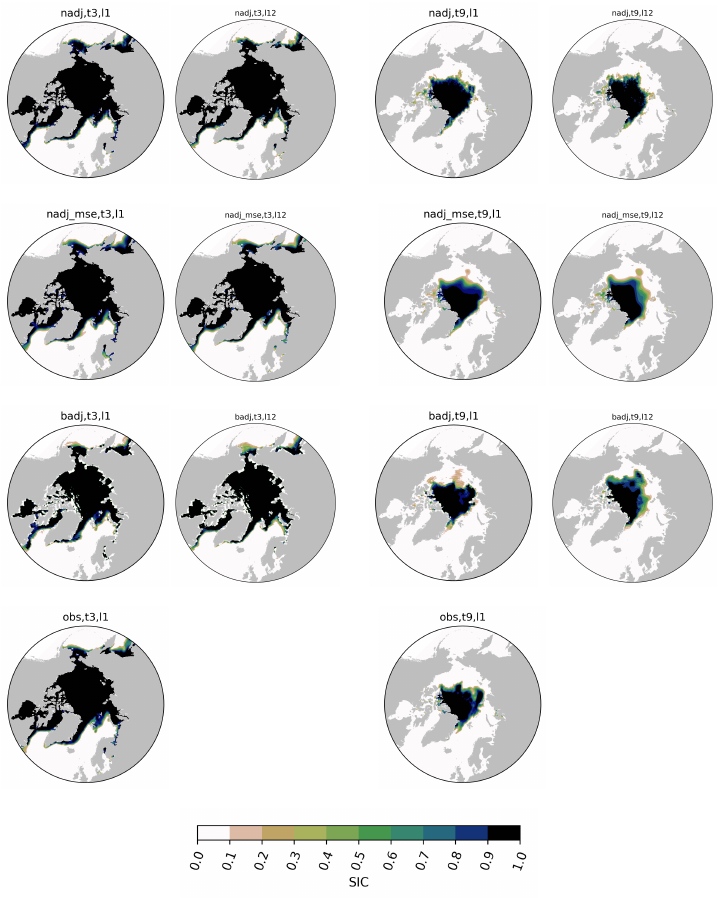}
\caption{Random ensemble member SIC maps for target time of $2023$ March maximum (t3 in first and second columns), and $2023$ September minimum (t9 in third and forth columns) on lead months 1 (l1), and 12 (l12). Rows are Nadj\_mse (first), Badj (second), Nadj (fourth) and observation (last).} 
\label{fig:figA1}
\end{figure}

\section{MBMC coefficients}
\label{sec:MBMC}

\noindent The scaling coefficients for the MBMC benchmark adjustment are computed as follows \citep{Dirkson2026}: \\ 

\begin{equation}
\alpha_{ml} = \frac{\sigma^{O}_{ml}}{\sigma^{H}_{ml}} \left[ \frac{(M-1)R_{ml}+
\sqrt{(M-1)^2R_{ml}^{2}+4M}}{2M}\right]
\label{eq:mbmc_alpha}
\end{equation}

\begin{equation}
\beta_{ml}=\left[\frac{\left(\sigma^{O}_{ml}\right)^2-\alpha_{ml} R_{ml}\sigma^{H}_{ml}\sigma^{O}_{ml}}{
\left(\sigma^{e}_{ml}\right)^2}\right]^{1/2}
\label{eq:mbmc_beta}
\end{equation}

\noindent where $M$ denotes the ensemble size, $m$ and $l$ are respectively the specified initial and lead month, $R_{ml}$ the correlation between ensemble mean hindcast and observation, $\sigma^O_{ml}$ and $\sigma^H_{ml}$ are respectively the observed and ensemble mean climatological standard deviations, and $\sigma^e_{ml}$ denotes the variance about the ensemble mean, averaged over initial years.\\

\newpage
\bibliographystyle{ametsocV6}
\bibliography{References.bib} 

\end{document}